\documentclass[10pt,twocolumn,letterpaper]{article}

\usepackage{cvpr}              

\usepackage{cuted} 

\usepackage{currfile} 

\usepackage{caption} 
\definecolor{cvprblue}{rgb}{0.21,0.49,0.74}
\usepackage[pagebackref,breaklinks,colorlinks,allcolors=cvprblue]{hyperref}
\newcommand{\methodname}{\textbf{LightHarness}}

\usepackage{multirow}
\usepackage{graphicx}
\usepackage{cuted}
\usepackage{caption}

\def\confName{CVPR}
\def\confYear{2027}

\title{Beyond Appearance: Evidence-Guided \\Programmatic 3D Scene Reconstruction with Agentic Lighting Reasoning}
\vspace{-10mm}
\author{
Zeting Luo \quad Yifan Li \quad Shuai Yang \\
Wangxuan Institute of Computer Technology, Peking University \\
{\tt\small 2400013131@stu.pku.edu.cn, liyifan02@stu.pku.edu.cn, williamyang@pku.edu.cn}\\
\url{https://logic-al-al.github.io/LightHarness/}
\vspace{-12mm}
}

\begin{document}
\maketitle
\begin{strip}
    \centering
    \includegraphics[width=\textwidth]{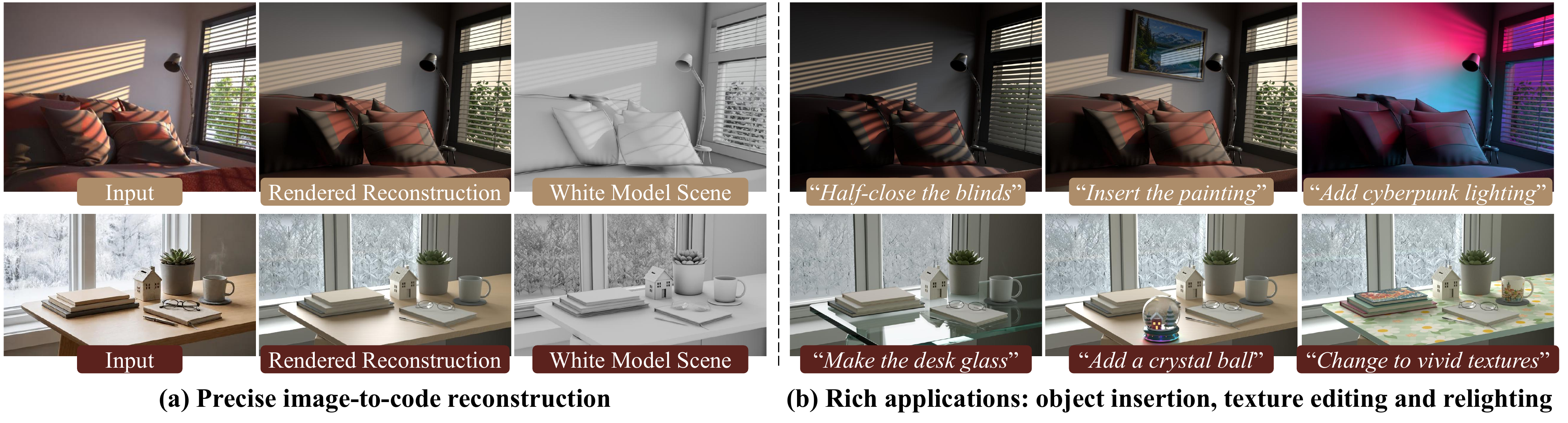}
    \captionof{figure}{
        \textbf{Lighting-aware image-to-code reconstruction.}
        Our framework reconstructs executable 3D scenes with explicit
        geometry, materials, and lighting, supporting diverse scene
        edits with realistic lighting interactions.
    }
    \label{fig:teaser}
    \vspace{-2mm}
\end{strip}

\begin{abstract}
Image-to-code methods for 3D scene reconstruction produce executable code that generates editable digital content and supports visual simulation.
However, existing methods primarily rely on appearance similarity, which provides limited guidance for distinguishing errors in geometry, materials, and illumination.
Light shapes scene appearance through its interactions with geometry and materials, influencing shading, cast shadows, and reflections.
We therefore present \methodname, a VLM-based agentic framework that incorporates explicit lighting reasoning into iterative scene reconstruction.
To be specific, we leverage complementary geometric, intrinsic, and illumination cues to jointly refine scene structure and appearance, and propose Lighting Reasoning Memory to maintain light-source hypotheses, which supports evidence, and revision history throughout iterations.
Together, they support reasoning about partially visible and unseen sources, with evidence-guided verification directing subsequent revisions.
Our framework reconstructs explicit scenes through coherent reasoning about geometry, materials, and illumination without additional training or differentiable rendering.
Extensive experiments demonstrate superior image-to-code reconstruction accuracy under challenging shadows and complex illumination, further supporting downstream scene editing, object insertion, and versatile relighting.
\end{abstract}    
\vspace{-4mm}
\section{Introduction}
\label{sec:introduction}

Reconstructing an executable, editable 3D scene from a single image connects visual understanding with digital scene creation.
Image-to-code reconstruction offers a promising direction: a scene program (\textit{e.g.}, Blender, Three.js) explicitly represents geometry, spatial layout, materials, cameras, and light sources, enabling novel-view rendering and subsequent scene editing~\cite{yin2026viga,he2026seig,wang2026threecos}.
Except for  spatial grounding that concerns object poses or location, \textbf{lighting} is also central to this process.
Shading, cast shadows, highlights and reflections arise from the complex interactions between illumination, geometry, and materials.
Accurate image-to-code reconstruction requires recovering object structure and layout together with the illumination that explains their observed appearance.

\begin{figure}[t]
    \centering
    \includegraphics[width=0.99\linewidth]{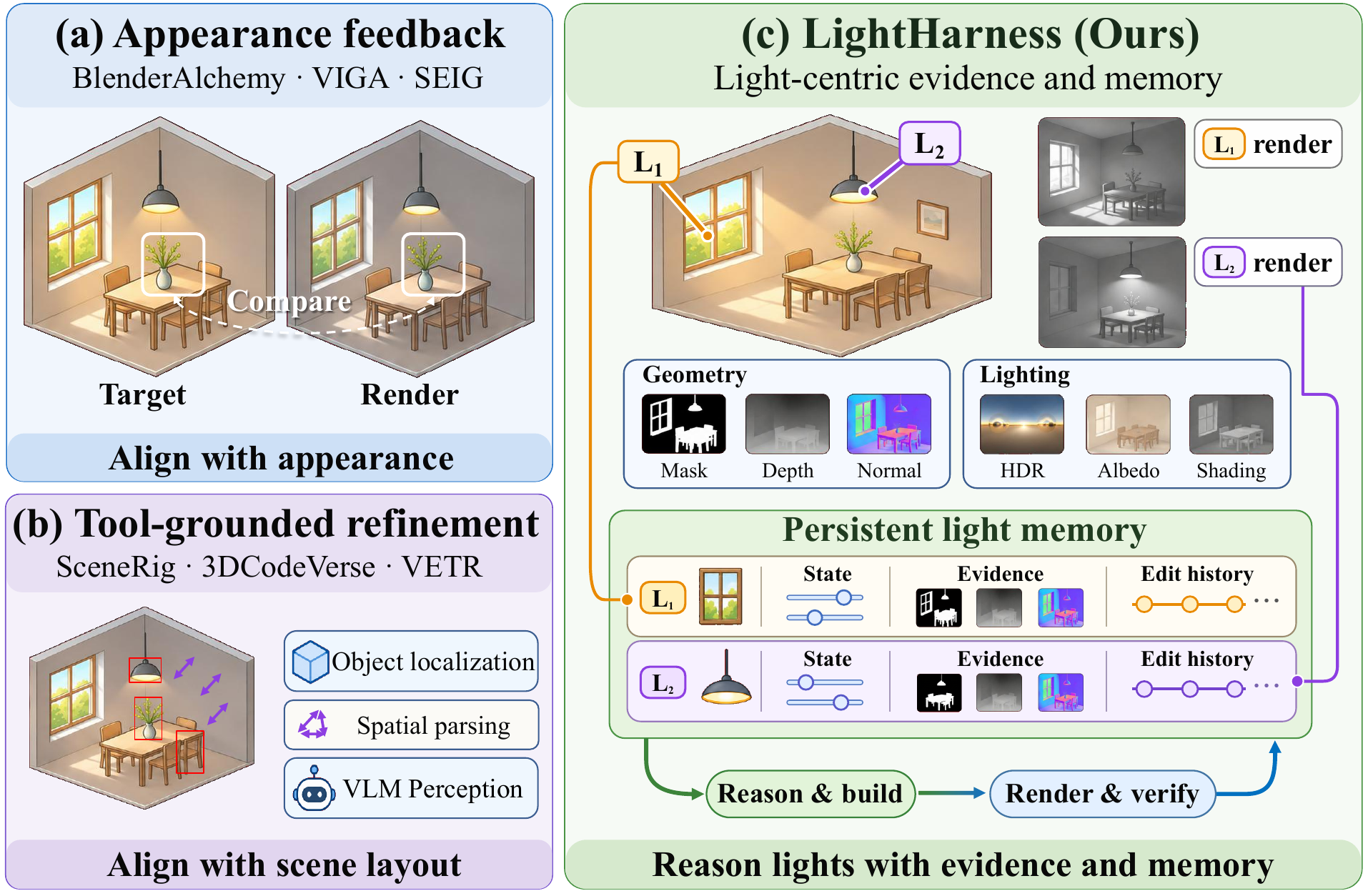}
    \caption{\textbf{Conceptual comparison of image-to-code paradigms.}
    \textbf{(a)} Appearance feedback emphasizes rendered--reference similarity.
    \textbf{(b)} Tool-grounded refinement uses perception and spatial analysis to improve layout.
    \textbf{(c)} LightHarness combines enriched evidence and persistent memory for explicit lighting reasoning throughout reconstruction.}
    \label{fig:concept_compair}
    \vspace{-5mm}
\end{figure}

Recent advances in vision-language models (VLMs)~\cite{openai2026gpt6astra,deepmind2026gemini4carbon} have enabled scene construction through executable programs, connecting visual and semantic reasoning with explicit control over scene components.
This capability supports a common iterative paradigm: an agent proposes or edits a scene program, inspects the rendered result, and revises its proposal based on visual feedback. 
SceneCraft~\cite{hu2024scenecraft} and BlenderAlchemy~\cite{huang2024blenderalchemy} demonstrate this paradigm to be effective in text-driven construction and appearance-guided editing, respectively. 
Extending it to reconstruction from visual input enjoys richer editing applications, and requires the VLMs to determine which scene configuration could explain the observed image. 
Recent systems formulate this problem through plan and observation management~\cite{yin2026viga}, staged refinement of scene attributes~\cite{he2026seig}, or explicit geometric, material, and light modeling~\cite{sun2026harmony,chen2026lumera} in synthetic game scenes. 

Across these approaches, rendered-reference comparisons provide a central feedback signal for scene revision.
However, such a discrepancy on appearance does not uniquely identify its underlying cause, because similar image differences can arise from geometry, reflectance, or illumination. 
For example, a dark region may arise from low surface reflectance, a surface facing away from light sources, or insufficient illumination from an unseen source.
Simply brightening the material can reduce a local image discrepancy, but leave the underlying illumination error unresolved, leading to an incorrect material estimation.
The challenge is therefore to connect observed lighting effects to their plausible geometric, material, and illumination explanations.
This motivates our central question: \textbf{How can we explicitly reason about scene geometry, materials, and light sources for more accurate image-to-code reconstruction beyond appearance?}

To mitigate this, we propose \methodname, a VLM-based agent framework that \textbf{incorporates explicit lighting modeling into iterative scene generation and verification with complementary evidence and reasoning}.
We first associate light-source hypotheses with evidence of their interactions with scene geometry and materials.
Specifically, we combine estimated HDR illumination~\cite{chen2026lumen}, depth and surface orientation~\cite{wang2025vggt,ke2025marigold}, intrinsic albedo and shading~\cite{ke2025marigold}.
These complementary sources of evidence relate appearance discrepancies to candidate revisions of the underlying scene factors, helping the \textit{generator agent} determine which geometry, material, or light parameters to update during reconstruction, and finally output an executable Blender program.
As for the \textit{verifier agent}, we equip it with a rubric scorer and a VLM. The rubric scorer assesses rendered-reference agreement based on predefined criteria, while the VLM uses these assessments, reference-image history revisions, and decomposed rendering evidence to diagnose and propose refinement updates. 

To preserve and refine illumination reasoning iteratively in the proposed agentic system, we design \textbf{Lighting Reasoning Memory} (LRM), shared by the generator and verifier agents.
LRM associates each hypothesized light source with a group of stable parameters, including identity, visibility, location, color, \textit{etc.}, establishing a clear correspondence among observations, editable Blender elements and the revision history.
\textbf{Notably}, we explicitly consider partially visible, occluded, and off-frame light sources as hypotheses, enabling a more complete model of scene illumination that accounts for both visible and inferred light sources. 
In each iteration, the generator uses this persistent lighting proxy to build scene programs and refine parameters, while the verifier examines it based on rendered evidence and recommends corrections.
By incorporating source-specific observations, decisions, and their supporting evidence in a synergistic framework, LRM provides persistent illumination-centric context for \methodname, enabling subsequent geometry, material, and lighting revisions to build coherently on historical judgments.

Extensive experiments demonstrate our competitive 3D image-to-code reconstruction accuracy.
We further verify our effectiveness on hard cases under complex lights and shadows, and showcase that our method can generalize to rich downstream applications, including image relighting, material/texture editing, and harmonious object insertion.
Our contributions can be summarized as follows:
\begin{itemize}
    \item We propose a VLM-based 3D image-to-code reconstruction framework that incorporates explicit lighting modeling to guide the recovery of geometry, materials, and light sources in an executable scene program.
    \item We introduce Lighting Reasoning Memory to preserve source identities, supporting evidence, and revision history across generation and verification, combining geometric and intrinsic priors with light-transport evidence to guide iterative scene refinement.
    \item We demonstrate competitive 3D image-to-code reconstruction accuracy and further illustrate the utility of the reconstructed scenes through downstream applications, including image relighting, enhancement and complex light understanding.
\end{itemize}

\section{Related Work}
\label{sec:related_work}

\noindent\textbf{Lighting Estimation.}
Inverse rendering~\cite{li2020inverse} aims to estimate scene geometry, reflectance, and illumination to explain observed image appearance and enable editing~\cite{li2022editing}.
PhotoScene~\cite{yeh2022photoscene} optimizes materials and lighting on approximately aligned CAD geometry, while Materialist~\cite{wang2026materialist} combines intrinsic prediction with differentiable material and environment-light refinement.
Multi-view methods further constrain material and illumination recovery through shared observations~\cite{wu2023fipt,lin2025iris}.
LUMEN~\cite{chen2026lumen} instead predicts local HDR illumination, providing a lighting prior rather than a complete scene program.
Generative relighting also incorporates physical guidance and local control through light-transport consistency~\cite{zhang2025iclight}, intrinsic conditioning and scribbles~\cite{choi2025scribblelight}, or estimated illumination priors~\cite{yen2026freelit}.
These methods provide complementary forms of lighting estimation and control, with different assumptions about available geometry and observations.
Our focus is to use such evidence to guide geometry, material, and light revisions during executable scene reconstruction, with relighting as a downstream application.

\noindent\textbf{Executable Scene Representations and Agentic Reconstruction.}
SceneCraft~\cite{hu2024scenecraft} translates scene descriptions and spatial constraints into Blender programs, while BlenderAlchemy~\cite{huang2024blenderalchemy} searches scene edits through rendered feedback.
For image-to-code reconstruction, VIGA~\cite{yin2026viga} combines generation and verification with multimodal memory, and SEIG~\cite{he2026seig} refines geometry, materials, composition, and lighting in stages.
HARMONY~\cite{sun2026harmony} and Lumera~\cite{chen2026lumera} incorporate geometric priors and explicit illumination, with Lumera retaining shared refinement history.
VETR~\cite{ma2026vetr} additionally uses execution observations to distinguish proposed edits from their realized effects.
Notably, Code Video Model~\cite{liang2026codevideomodel} combines executable codes with frontier generative models, achieving versatile controllable video generation.
Although these systems already incorporate multiple forms of scene evidence and iterative feedback, our focus is to maintain source-specific lighting hypotheses and connect memorable evolution to geometric, intrinsic, and decomposed rendering evidence throughout reconstruction.

\section{Method}
\label{sec:method}

\begin{figure*}[t]
\centering
\includegraphics[width=\linewidth]{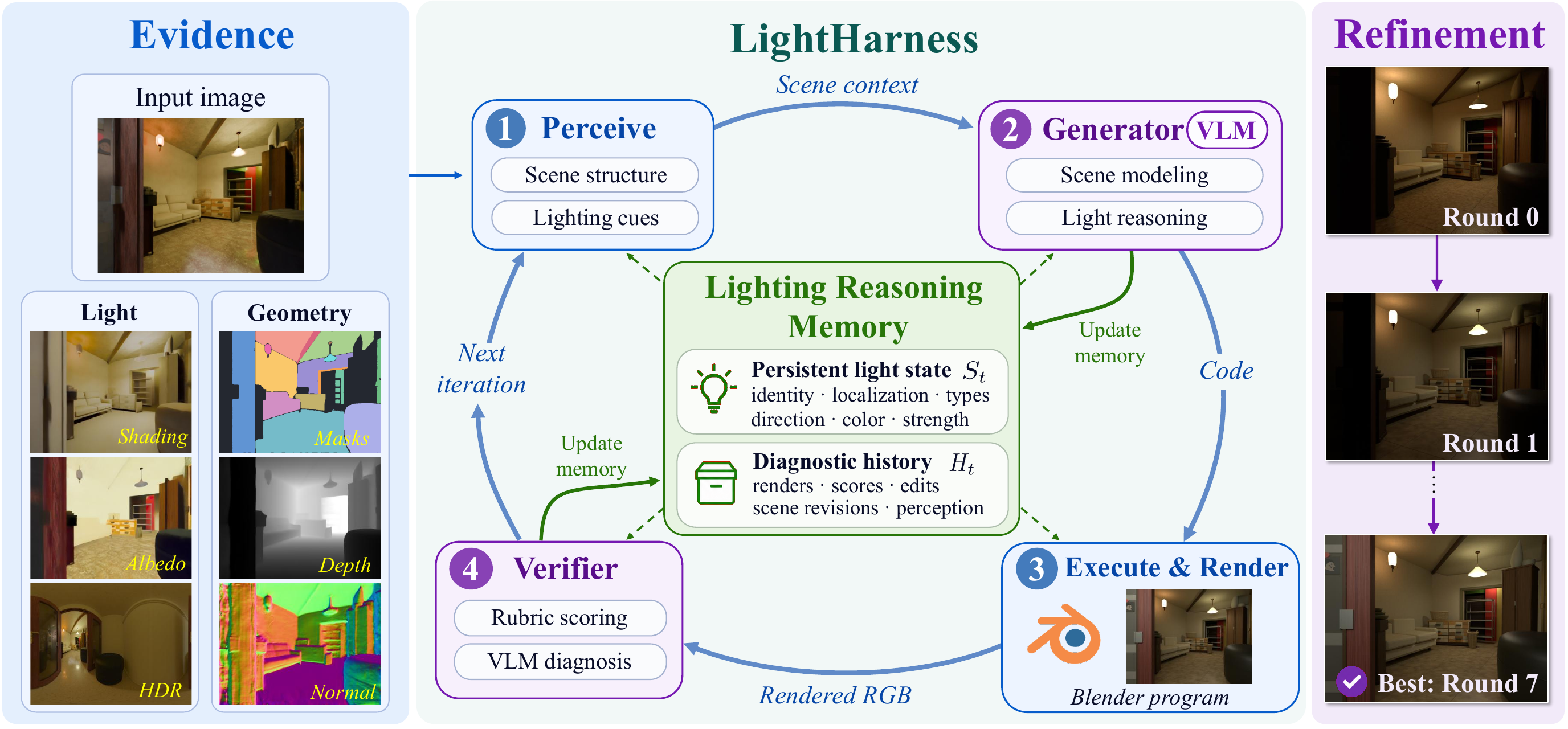}
\caption{\textbf{Overall pipeline of LightHarness.}
Geometric and lighting evidence guides a closed loop of perception, VLM-based generation, execution and rendering, and verification.
Rubric scoring and VLM diagnosis guide joint scene refinement, while Lighting Reasoning Memory preserves source states and diagnostic history across iterations.}
\label{fig:main_pipeline}
\vspace{-3mm}
\end{figure*}

\subsection{Motivation and Overview}
\label{sec:method_overview}

Image-to-code reconstruction represents geometry, materials, and lighting in an executable program, providing an interpretable proxy for rendering, scene manipulation, and downstream applications.
Existing methods primarily exploit this problem through scene composition, spatial constraints, and interactive environments~\cite{hu2024scenecraft,luo2026scenerig,gao2026threecodeverse}.
However, faithful reconstruction requires understanding how scene components interact with \textbf{light} to produce the observed appearance.
Inferring lighting from incomplete observations is ill-posed because different scene configurations can produce similar images.
A dark wall may indicate low reflectance, occlusion, or a surface facing away from illumination.
Misinterpreting these effects can lead to incorrect materials or geometry, hampering reconstruction fidelity and subsequent editing.
We therefore treat lighting reasoning as a crucial part of image-to-code reconstruction.

As shown in Fig.~\ref{fig:main_pipeline}, \methodname\ formulates a $T$-round loop which follows \emph{Perceive $\rightarrow$ Generator $\rightarrow$ Execute \& Render $\rightarrow$ Verifier} and combines visual evidence $E$ extracted from image $I$ with Lighting Reasoning Memory (LRM) $M_t$.
Perception forms the scene context, which guides the Generator in producing code $C_t$ and light-source hypotheses.
Following rendering, rubric scoring and VLM diagnosis assess the reconstruction and guide subsequent revisions.
Generated scene states and verification findings update LRM, preserving evidence and reasoning across rounds.
The best valid candidate is selected by its rubric score, yielding jointly refined geometry, materials, and lighting for a wide range of downstream applications.

\subsection{Enriched Evidence for Lighting Reasoning}
\label{sec:evidence}

Lighting reasoning requires both the spatial structure governing light transport and evidence of its effects on surfaces.
We extract complementary geometric, intrinsic, and lighting cues to support these judgments.

\noindent\textbf{Geometric evidence.}
SAM~3~\cite{carion2026sam3} provides fine-grained semantic object masks, while
VGGT~\cite{wang2025vggt} estimates relative depth and cameras to establish a common reconstruction coordinate system.
For planar structures, we back-project reliable masked depth pixels and fit planes through robust estimation using least-squares refinement.
Each plane provides a surface point and normal for reasoning about position, orientation, and illumination.
Non-planar objects retain masks and depth cues without planar fitting.

\noindent\textbf{Intrinsic and lighting evidence.}
Marigold~\cite{ke2025marigold} estimates albedo and diffuse shading to help distinguish reflectance from light effects.
LUMEN~\cite{chen2026lumen} offers an HDR image suggesting light-source directions and chromatic properties.
Evidence is extracted once and remains fixed, while its interpretation and reasoning evolve with rendering feedback.

\subsection{Lighting Reasoning Memory}
\label{sec:lighting_memory}

Scene revisions can change a light source's contribution even when its parameters remain unchanged.
Lighting reasoning must therefore retain both current source states and the evidence underlying previous decisions.
Therefore, we propose shared Lighting Reasoning Memory (LRM):
\begin{equation}
    M_t = (S_t,H_t),
\end{equation}
where $S_t$ contains light-source states and $H_t$ records diagnostic history.
We connect the stable light-source identities with corresponding Blender elements, where each state describes light visibility, type, position, orientation, intensity and color.
VLM visual reasoning~\cite{openai2026gpt6astra,deepmind2026gemini4carbon}, guided by enriched evidence, also supports hypotheses of partially visible, occluded, and off-frame light sources.

The history $H_t$ associates sources with previous assessments, proposed revisions, implemented changes, and observed outcomes.
It further relates multiple sources through semantic correspondence to support a coherent interpretation of scene illumination.
Together, $S_t$ and $H_t$ guide subsequent geometry, material, and lighting reasoning.
Their update is described as below.

\subsection{Evidence-Guided Verification}
\label{sec:score}

The Verifier consists a deterministic rubric scorer and a VLM diagnosis to assess reconstruction quality and guide scene refinement.
The scorer measures agreement with the reference image and enriched evidence, while the VLM interprets these measurements alongside rendered outputs and LRM to identify plausible causes of discrepancies.

\noindent\textbf{Rubric-based assessment.}
The rubric is six-fold as below.
(1) \textit{Object-mask error} measures projected-region disagreement.
(2) \textit{Relative-depth error} compares reliable overlapping regions after positive scale alignment and penalizes missing coverage.
(3) \textit{Orientation error} compares depth-derived plane normals while accounting for sign ambiguity.
(4-5) \textit{Albedo and diffuse-shading errors} are measured in trusted diffuse regions in linear color space.
Shading uses one positive alignment scale shared across channels to preserve chromatic differences.
(6) \textit{RGB error} evaluates overall appearance.
We aggregate these errors into scores $q_t$:
\begin{equation}
    q_t = \frac{\sum_{k\in\mathcal{K}} w_k e_{t,k}}
                   {\sum_{k\in\mathcal{K}} w_k},
    \label{eq}
\end{equation}
where $\mathcal{K}$ indexes mask, depth, orientation, albedo, shading, and RGB errors.
Corresponding weights $w_k$ remain fixed across rounds at $(2,1,0.5,0.5,1,1)$, with sensitivity examined through ablations provided in the Supp.
Lower scores indicate better agreement and guide candidate selection without gradient-based optimization.

\noindent\textbf{Evidence-guided diagnosis.}
Beyond scalar assessment, the VLM combines component errors, rendered passes, source contributions, and LRM to recommend geometric, material, or lighting revisions.
For an underlit wall, it examines orientation, occlusion, albedo, and relevant source contributions.
Geometric discrepancies motivate structural revisions, whereas consistent geometry and albedo direct attention to lighting.
Each recommendation specifies the discrepancy, proposed revision, and supporting evidence, providing guidance for the subsequent reconstruction round.

\subsection{Memory-assisted Scene Reconstruction Loop}
\label{sec:scene_generation}

With enriched evidence, LRM, and the Verifier defined, we integrate them into a closed loop of perception, generation, execution/rendering, and verification.
At round $t$, the perception module combines the input image $I$, fixed evidence $E$, and memory $M_t$ into a scene context comprising structural cues, lighting hypotheses, and unresolved discrepancies.
Conditioned on this context and preceding scene code, the Generator jointly reasons about structure and illumination to produce $t$-step code $C_t$ and a light source state update $\Delta S_t$.
After checking state consistency and source correspondence, we obtain
\begin{equation}
    (C_t, \Delta S_t) =\text{VLM}(I,E,M_t, C_{t-1}), \quad\widetilde S_t = \operatorname{Apply}(S_t,\Delta S_t).
\end{equation}

The generated geometry and materials are combined with $\widetilde S_t$ under fixed camera and rendering settings.
Blender $\mathcal{R}$ then renders the candidate scene as

\begin{equation}
    Y_t = \mathcal{R}(C_t,\widetilde S_t),
\end{equation}
including RGB, masks, depth, surface information, albedo, diffuse shading, and individual light sources and world-illumination contributions.
Masks and depth show scene layout and occlusion. Albedo and shading help separate material color from lighting, while light source attribution indicates which region it affects.
The Verifier assesses these outputs against the reference evidence, producing the rubric score $q_t$ and evidence-guided recommendations.

Following each valid round, we reorganize the light source state and append a record $h_t$ containing the diagnosis, implemented changes, scores, and references to archived outputs:
\vspace{-3mm}
\begin{equation}
    M_{t+1} = (\widetilde S_t,\ H_t\mathbin{\|}h_t),
\end{equation}
where $\|$ denotes history concatenation.
The updated memory guides perception and generation in the next round, allowing subsequent reasoning to build on previous evidence and outcomes.

Since revisions may improve some factors while degrading others, we select
\vspace{-3mm}
\begin{equation}
    t^* = \arg\min_{t\in\mathcal{T}_{\mathrm{valid}}} q_t
\end{equation}
over completed valid rounds after $T$ iterations.
We return $(C_{t^*},\widetilde S_{t^*})$ together, preserving the correspondence between the selected executable scene and its lighting state.
This closed loop connects evidence, diagnosis, and persistent memory to support coherent refinement of geometry, materials, and illumination.

\vspace{-3mm}
\section{Experiments}
\label{sec:experiments}

Our experiments compare reconstruction quality with recently blooming image-to-code methods, examine the roles of visual evidence and Lighting Reasoning Memory, and assess generalizable downstream application and hard cases.

\subsection{Implementation Details}
\label{sec:experimental_setup}

\paragraph{Framework development.}
We load external modules (\textit{i.e.}, SAM3, VGGT, Marigold, LUMEN) on two NVIDIA 4090-48G GPUs, and use GPT-6-Astra~\cite{openai2026gpt6astra} as the Generator/Verifier VLM via API. All reconstructions are done with 10 rounds.
We also examine our framework under different VLMs in the Supplementary materials.

\vspace{-4mm}
\paragraph{Data.} 
We evaluate on 32 images drawn from diverse benchmarks from robotics, relighting and image enhancement, comprising 10 from Hypersim~\cite{hypersim},
4 from LuxRemix~\cite{lux}, 10 from BAID~\cite{baid},
7 from MIT-Adobe FiveK~\cite{fivek}, and 1 from RoboDojo~\cite{chen2026robodojo} with complex lighting conditions, diverse geometry and materials. The whole testing set contains 17 real photographs and 15 synthetic images, covering 22 indoor and 10 outdoor scenes.

\vspace{-4mm}
\paragraph{Metrics.}
We evaluate source-view consistency using PSNR~\cite{wang2009mse},
SSIM~\cite{wang2004ssim}, LPIPS~\cite{zhang2018lpips},
and DreamSim~\cite{fu2023dreamsim}.
Following SceneRig~\cite{luo2026scenerig}, we assess 3D reconstruction quality from the source view and nearby views at $\pm15^\circ$.
By orbiting the camera around the world vertical axis through a pivot located four scene units along the original camera's viewing direction, we use GPT-6-Astra as a judge to compute the mean-centered Bradley-Terry scores among these rendered pairs, with higher scores indicating
stronger relative preference.

\vspace{-5mm}
\paragraph{Compared methods.}
We compare with recent SoTA image-to-code reconstruction methods, including VIGA~\cite{yin2026viga}, 3DCodeVerse~\cite{gao2026threecodeverse}, SceneRig~\cite{luo2026scenerig}.
All these methods use GPT-6-Astra as the backbone VLM with the same loop iterations and API. We additionally compare GPT-6-Sol with Codex~\cite{codex} harness.

\begin{figure*}[t]
    \centering
    \includegraphics[width=0.99\linewidth]{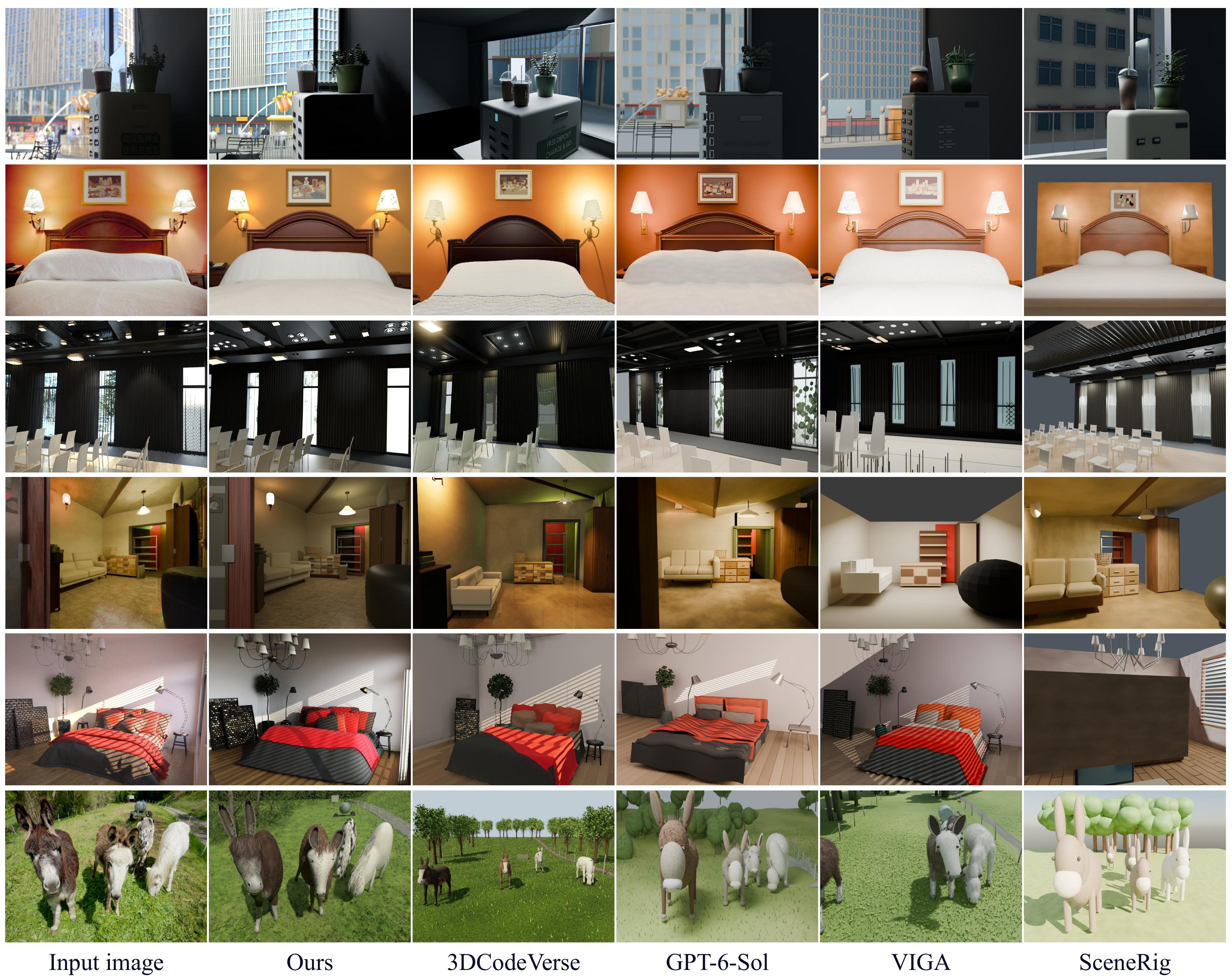}
    \caption{\textbf{Qualitative comparison between other SoTA image-to-code baselines.} Across diverse indoor and outdoor scenes, our \methodname\ better preserves object arrangement and characteristic lighting effects, including strong back-lighting, localized illumination, and structured cast shadows, achieving more faithful reconstructions of scene structure and appearance.}
    \label{fig:qualitative_v2}
    \vspace{-5mm}
\end{figure*}

\subsection{Image-to-Code Reconstruction}
\label{sec:main_comparison}

\paragraph{Qualitative results.}
\label{sec:qualitative_results}
Fig.~\ref{fig:qualitative_v2} compares reconstructions across diverse geometry and lighting conditions.
Our method better captures window–foreground contrast, localized illumination around bedside lamps, and combined window and ceiling lighting in the hall.
It also preserves furniture arrangement and foreground animal composition, while competing methods exhibit layout distortions, simplified geometry, or mismatched illumination, supporting joint reasoning about scene structure, materials, and lighting.
\textbf{Notably}, our method infers unseen light sources and structures through their illumination effects.
In the fifth row, it reproduces striped shadows across the bedroom wall and bedding while preserving their spatial relationships, whereas competing methods miss these shadows or misinterpret the layout.
The cast shadows in the last row further illustrate lighting reasoning \textbf{beyond appearance}, enabling more faithful image-to-code reconstruction.
We present more visualized results under nearby views in the Supp.

\vspace{-3mm}
\paragraph{Hard case verification.}
\begin{figure}[t]
    \centering
    \includegraphics[width=0.8\linewidth]{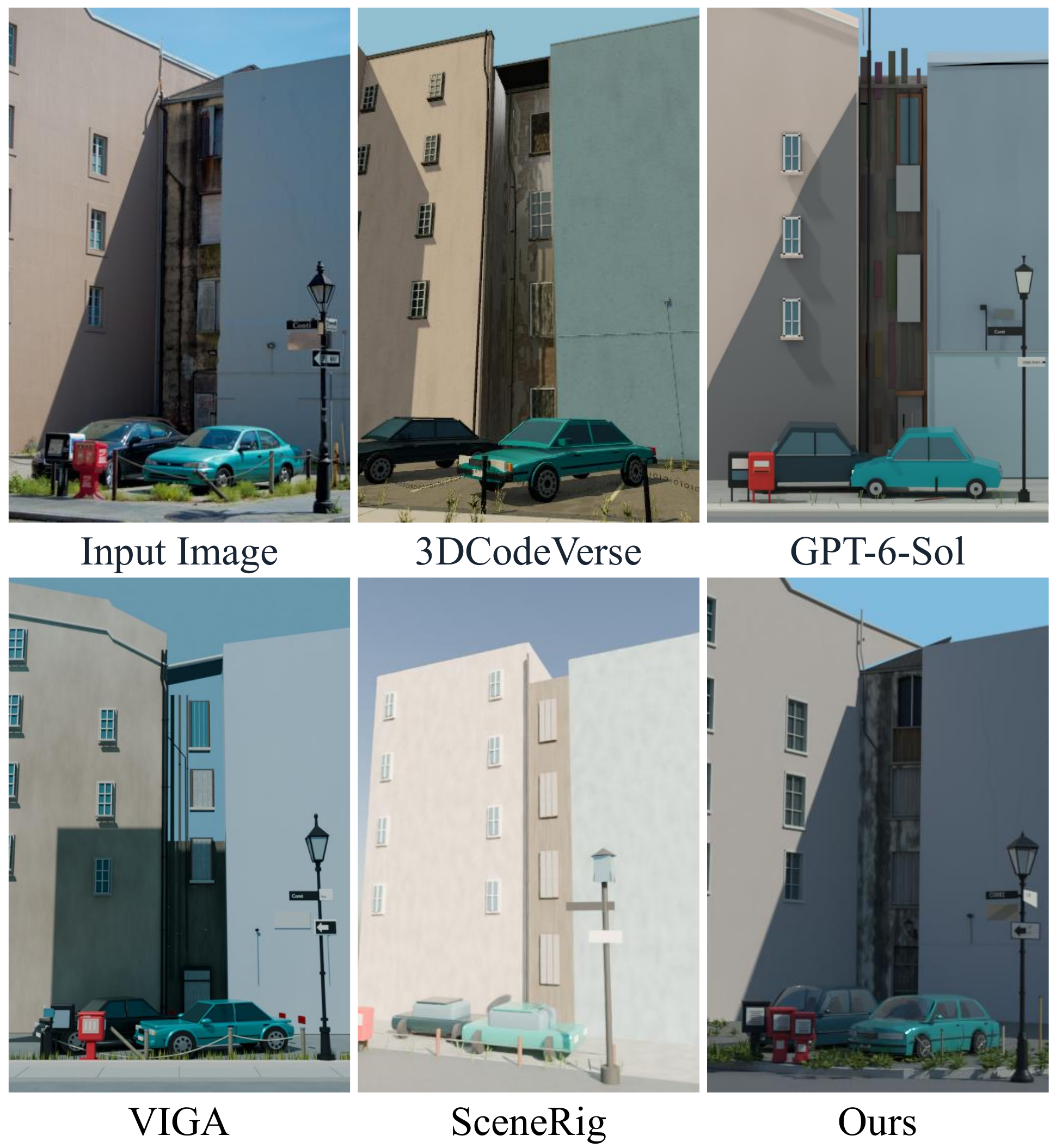}
    \vspace{-3mm}
    \caption{\textbf{Hard case verification.}
    Our method more faithfully reconstructs the diagonal shadow and surrounding building structure relative to other methods.}
    \vspace{-3mm}
    \label{fig:hard_case}
\end{figure}

Fig.~\ref{fig:hard_case} compares reconstructions of a scene with a prominent diagonal cast shadow.
While competing methods distort or omit this shadow, \methodname\ better reproduces its spatial extent while preserving the surrounding building structure.
This result supports the value of enriched evidence and persistent lighting memory in relating shadow patterns to scene geometry and illumination, enabling more faithful image-to-code reconstruction under challenging lighting.

\vspace{-6mm}
\paragraph{Quantitative results.}

Table~\ref{tab:main_reconstruction} shows that our method achieves the best PSNR, LPIPS, and DreamSim among the compared methods.
Relative to the recent SoTA method, VIGA, \methodname\ improves PSNR by 0.75\,dB and reduces DreamSim by 10.7\%, indicating our superiority in both pixel-level fidelity and perceptual similarity.
The VLM-based evaluation and user preference further demonstrate our advantages at the scene level, ranking it first in all criteria covering completeness, layout, and appearance.
These consistent preferences support the effectiveness of our framework in reconstructing scene composition and visual appearance.

\vspace{-6mm}
\paragraph{Inference cost.} The latency mainly includes evidence extraction, GPT API and Blender rendering in the generation-verification loop. For a 768$\times$576 image, these three parts cost 4.5 minutes, 118.8 minutes, and 1.1 minutes respectively. The latency is comparable with other baselines (VIGA~\cite{yin2026viga} for 71.2 minutes and 3DCodeVerse~\cite{gao2026threecodeverse} for 126.3 minutes). The critical bottleneck is API invocation, while other parts are relatively efficient with less than 6 minutes on two 4090-48 GB GPUs.

\vspace{-5mm}
\paragraph{User Study.} We invite 29 participants to rank the reconstructed scenes according to \textbf{(1)} scene structure fidelity and \textbf{(2)} lighting appearance fidelity, using the reference images as guidance. We achieve the first-place preference rate on both criteria, indicating stronger human preference for its reconstruction of scene structure and illumination.

\begin{table*}[t]
\centering
\caption{\textbf{Quantitative comparison of image-to-code reconstruction.} We report four appearance consistency metrics covering PSNR, SSIM, LPIPS and DreamSim, and measure novel-view consistency by a mean-centered Bradley-Terry (BT) score which is estimated from pairwise judgments using an additional $\pm15^{\circ}$ view following SceneRig~\cite{luo2026scenerig}.
Bold and underline indicate the best and second-best.}
\label{tab:main_reconstruction}
\small
\setlength{\tabcolsep}{4pt}
\renewcommand{\arraystretch}{0.95}
\begin{tabular}{lccccccccc}
\toprule
 & \multicolumn{4}{c}{Source-view reconstruction} & \multicolumn{3}{c}{VLM evaluation (BT $\uparrow$)} & \multicolumn{2}{c}{User Preference $\uparrow$}\\
\cmidrule(lr){2-5}\cmidrule(lr){6-8}\cmidrule(lr){9-10}
Method & PSNR $\uparrow$ & SSIM $\uparrow$ & LPIPS $\downarrow$ & DreamSim $\downarrow$ & Completeness & Layout & Appearance & Light & Structure \\
\midrule
3DCodeVerse~\cite{gao2026threecodeverse} & $10.21$ & $0.428$ & $0.677$ & $0.363$ & $-0.367$ & $-0.830$ & $-0.594$ & $6.5\%$ & $3.8\%$ \\
GPT-6 Sol & $13.03$ & $\underline{0.503}$ & $0.548$ & $0.318$ & $-0.325$ & $-0.284$ & $-0.262$ & $6.9\%$ & $2.5\%$ \\
VIGA~\cite{yin2026viga} & $\underline{14.28}$ & $\mathbf{0.525}$ & $\underline{0.446}$ & $\underline{0.242}$ & $\underline{0.524}$ & $\underline{0.537}$ & $\underline{0.700}$ & $\underline{21.4\%}$ & $\underline{17.8\%}$ \\
SceneRig~\cite{luo2026scenerig} & $11.86$ & $0.455$ & $0.665$ & $0.420$ & $-1.092$ & $-1.043$ & $-1.232$ & $3.3\%$ & $1.3\%$ \\
\midrule
\methodname\ (Ours) & $\mathbf{15.03}$ & $0.486$ & $\mathbf{0.437}$ & $\mathbf{0.216}$ & $\mathbf{1.259}$ & $\mathbf{1.620}$ & $\mathbf{1.388}$ & $\mathbf{61.9\%}$ & $\mathbf{74.6\%}$ \\
\bottomrule
\end{tabular}
\vspace{-4mm}
\end{table*}

\subsection{Ablation Studies}

\label{sec:ablation}

\begin{figure}
    \centering
    \includegraphics[width=\linewidth]{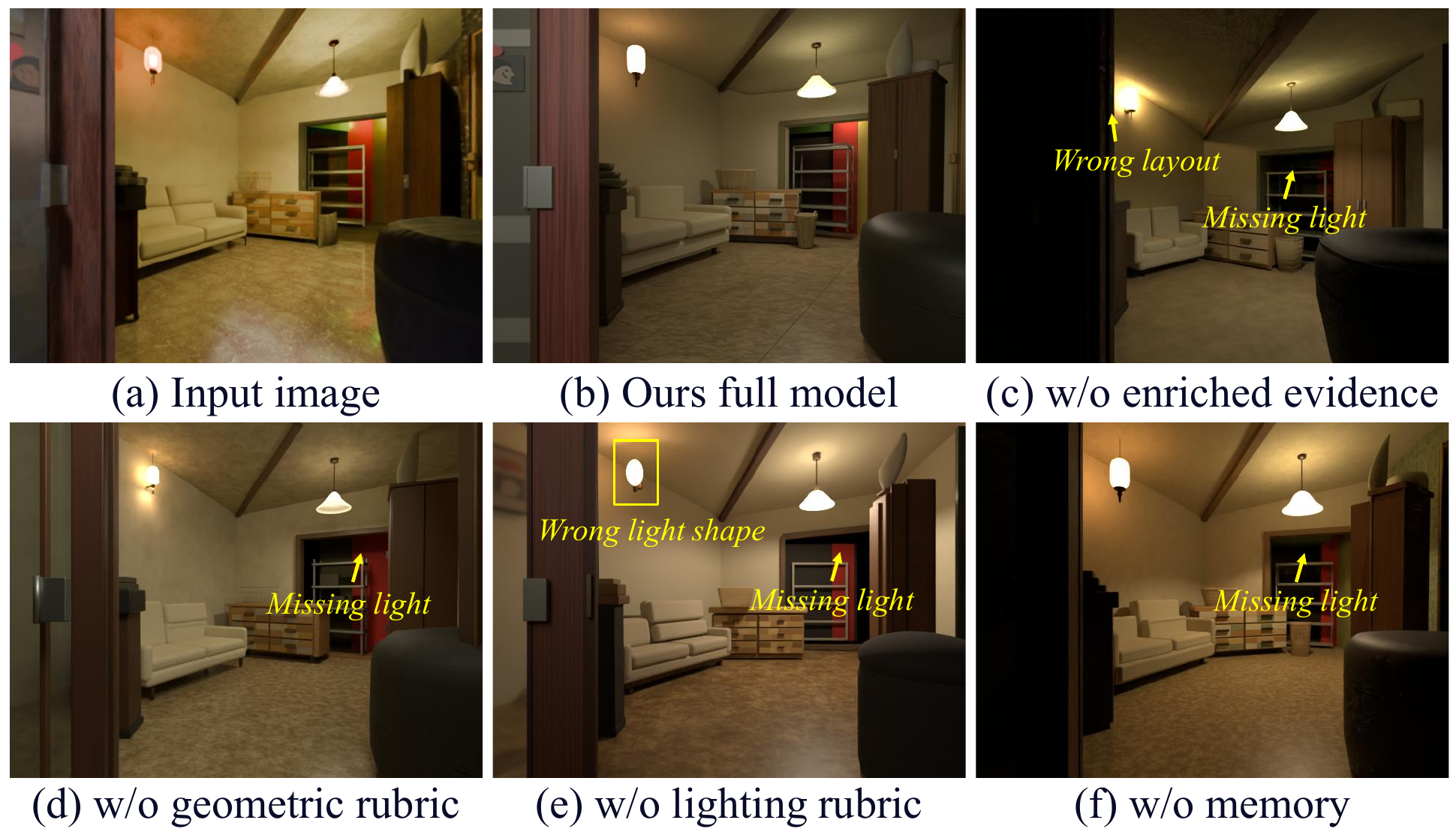}
    \vspace{-6mm}
    \caption{\textbf{Qualitative ablation of key designs.} Our full framework reproduces scene layout and illumination more precisely, while removing individual components introduces errors in basic scene layout (c), light placement (c-f), and light source shape (e).}
    \vspace{-5mm}
    \label{fig:ablation}
\end{figure}

\begin{table}[t]
\centering
\caption{\textbf{Quantitative ablation experiment.}
We evaluate on a 16-image subset for efficiency. Our full model achieves the best. BT scores are re-normalized here.}
\vspace{-3mm}
\label{tab:ablation}
\footnotesize
\setlength{\tabcolsep}{2.5pt}
\renewcommand{\arraystretch}{0.95}
\begin{tabular*}{\columnwidth}{@{\extracolsep{\fill}}lcccc@{}}
\toprule
Variant
& PSNR$\uparrow$
& DreamSim$\downarrow$
& \begin{tabular}[c]{@{}c@{}}BT$\uparrow$\\Compl.\end{tabular}
& \begin{tabular}[c]{@{}c@{}}BT$\uparrow$\\Appear.\end{tabular} \\
\midrule
w/o enriched evidence & $10.05$ & $0.351$ & $-0.4864$ & $-1.1035$ \\
w/o geometric rubric & $13.56$ & $0.278$ & $-0.7148$ & $-0.2589$ \\
w/o lighting rubric  & $13.70$ & $0.239$ & $0.3230$ & $0.5107$ \\
w/o LRM              & $12.81$ & $0.270$ & $-0.0078$ & $-0.0724$ \\
\midrule
\textbf{Full model}
& $\mathbf{14.24}$
& $\mathbf{0.213}$
& $\mathbf{0.8861}$
& $\mathbf{0.9241}$ \\
\bottomrule
\end{tabular*}
\vspace{-3mm}
\end{table}

We conduct quantitative and qualitative ablation experiments to demonstrate the effectiveness of each design in our framework in Fig.~\ref{fig:ablation} and Table~\ref{tab:ablation}.

\vspace{-5mm}
\paragraph{w/o enriched evidence.}
We disable the light and geometric evidence in Sec.~\ref{sec:evidence}, observing significant reconstruction error. As shown in Fig.~\ref{fig:ablation}~(c), the agent misjudges the wall light's distance from the surface and misses background illumination.

\vspace{-5mm}
\paragraph{w/o geometric rubric.}
Discarding explicit constraints on geometry features leads to incomplete scene light reconstruction, as shown in Fig.~\ref{fig:ablation}~(d).

\vspace{-5mm}
\paragraph{w/o lighting rubric.}
Discarding lighting measurement produces wrong light source shapes and mismatched local illumination strength as shown in Fig.~\ref{fig:ablation}~(e).

\vspace{-5mm}
\paragraph{w/o lighting reasoning memory.}
We compare our full model with a variant that retains only current scene code and parameter values.
Without memory, background illumination is missed as shown in Fig.~\ref{fig:ablation}~(f), highlighting the value of explicitly organizing light source hypotheses and diagnostic history across iterations.

\subsection{Generalizable Applications}
\label{sec:applications}
Building upon the precise executable scenes we reconstruct, our framework supports diverse downstream applications, including scene texture editing, harmonious object insertion, versatile relighting, and degraded illumination restoration.
We also demonstrate that our rendered edits can combine with the frontier generative model, GPT-Image~\cite{openai2025gptimage} to produce more physically realistic results.
Note that simply applying generative models is hard to achieve precise physical modeling like lighting interaction with complex shadows and textures. Our integration with GPT-Image enables advanced editing ability.
\begin{figure}
    \centering
    \includegraphics[width=0.99\linewidth]{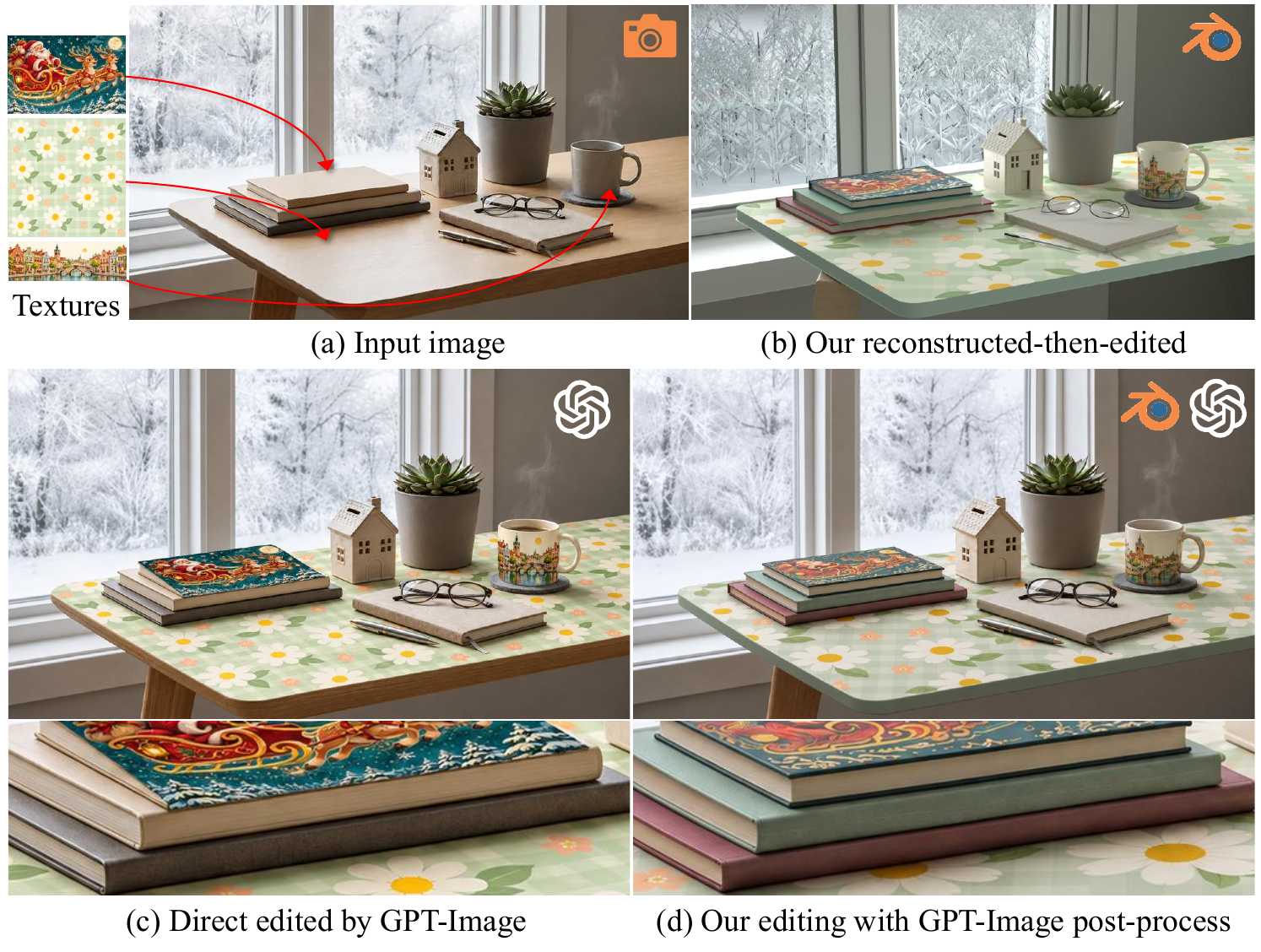}
    \vspace{-3mm}
    \caption{\textbf{Application of texture editing.} Our method supports texture replacement with multiple reference images with precise and executable code reconstruction, while directly applying even front-tier generative models leads to inaccurate structure.}
    \label{fig:app_texture}
    \vspace{-3mm}
\end{figure}

\vspace{-4mm}
\paragraph{Material and Texture Editing.}
Our method can perform exact texture editing with multiple references in Fig.~\ref{fig:app_texture}.

\begin{figure}
    \centering
    \includegraphics[width=0.99\linewidth]{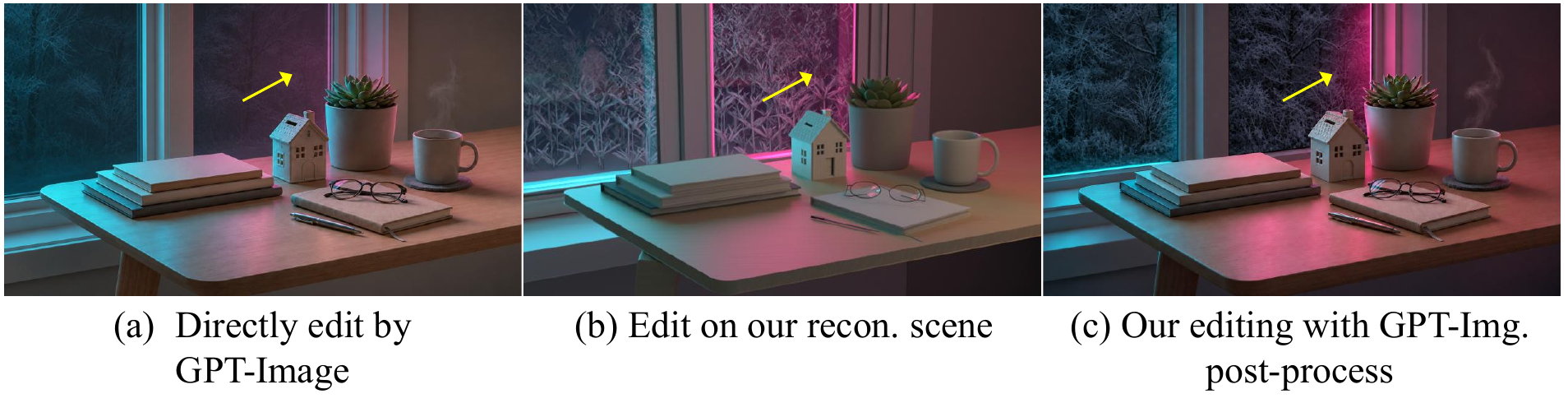}
    \vspace{-3mm}
    \caption{\textbf{Application of relighting with realistic reflection.}}
    \label{fig:app_relight}\vspace{-2mm}
\end{figure}

\begin{figure}
    \centering
    \includegraphics[width=0.99\linewidth]{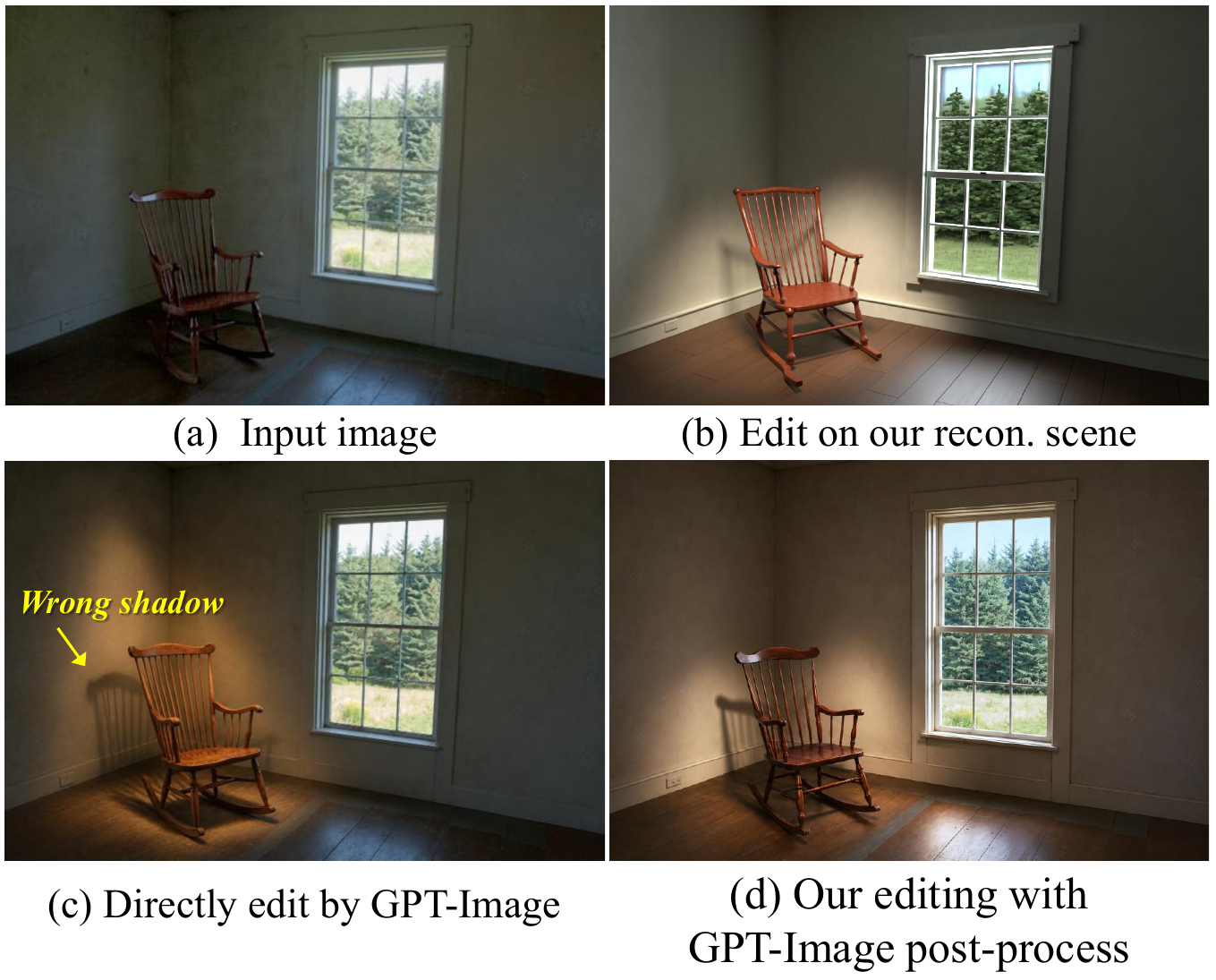}\vspace{-3mm}
    \caption{\textbf{Application of relighting with realistic shadow.}}\vspace{-7mm}
    \label{fig:app_relight_2}
\end{figure}

\vspace{-3mm}
\paragraph{Versatile relighting with flexible control.}
Our method achieves superior light control relative to simply applying GPT-Image, as shown in Fig.~\ref{fig:app_relight} and Fig.~\ref{fig:app_relight_2}.

\begin{figure}
    \centering
    \includegraphics[width=\linewidth]{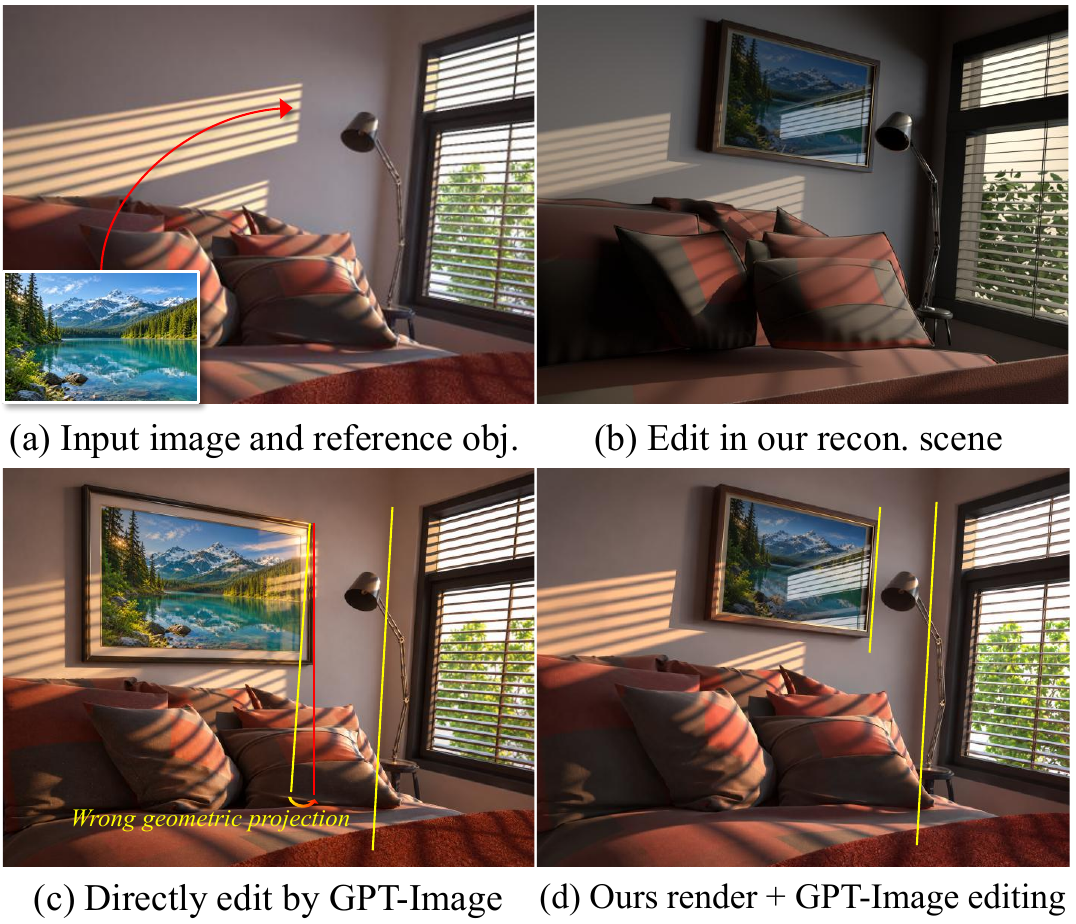}
    \vspace{-4mm}
    \caption{\textbf{Application of harmonious object insertion.} Our method \textbf{(b, d) }enables physically realistic lighting, reflection, occlusion and geometric projection, while \textbf{(c)} generative model-only method struggles with these real-world principles.}
    \label{fig:app_object_insert}
    \vspace{-2mm}
\end{figure}

\vspace{-3mm}
\paragraph{Harmonious object insertion.}
Our framework enables physically realistic object insertion as shown in Fig.~\ref{fig:app_object_insert}.

\vspace{-3mm}
\section{Conclusion, Limitation and Future Work}
We propose LightHarness, a VLM-based agentic framework that integrates explicit lighting reasoning and persistent light-source memory into image-to-code 3D reconstruction.
Complementary geometric, intrinsic, and illumination evidence guides reconstruction, while rubric-based assessment and VLM diagnosis support iterative refinement.
Lighting reasoning memory preserves source identities, evidence, semantic relationships, and revision history, enabling coherent reasoning across rounds.
Quantitative and qualitative results demonstrate our improved reconstruction fidelity under challenging shadows and complex illumination compared to recent SoTA methods.
Based on the accurate programmatic scene reconstruction, our framework enables rich applications, including relighting, harmonious object insertion, material and texture editing with plausible physical modeling.

\begin{figure}
    \centering
    \includegraphics[width=\linewidth]{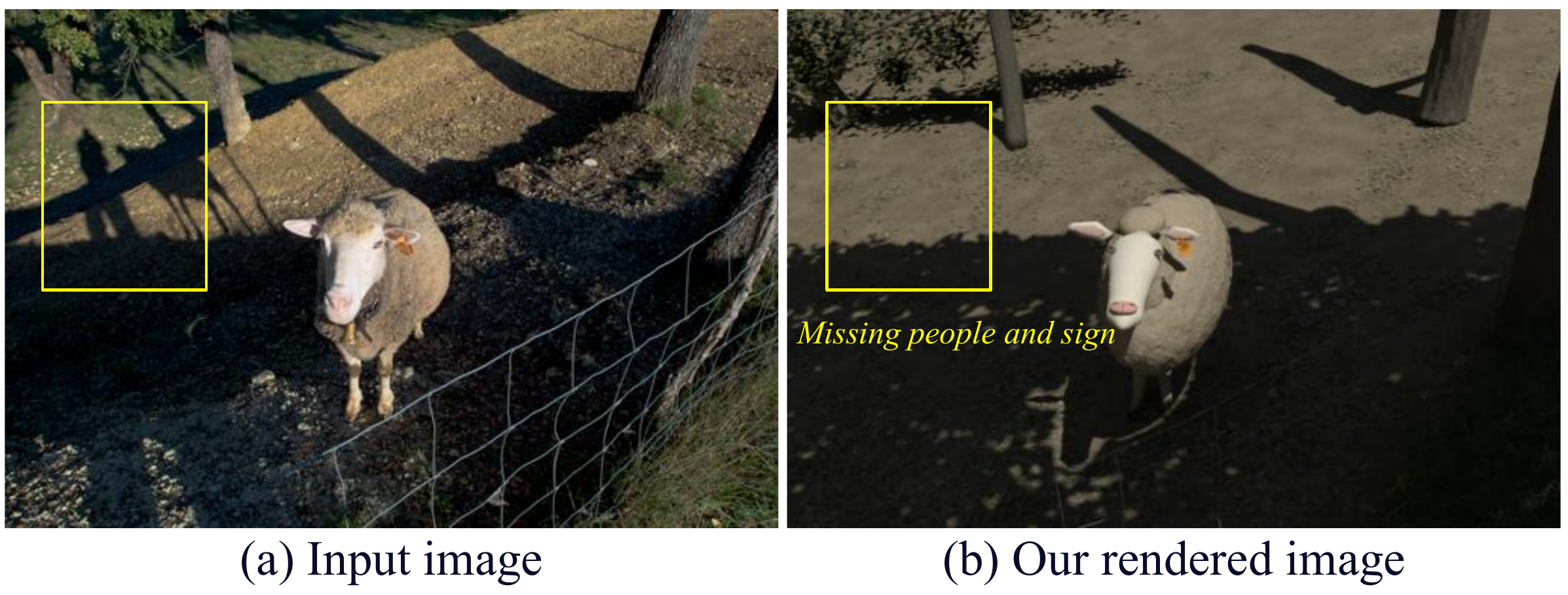}
    \vspace{-3mm}
    \caption{\textbf{Limitation on complex invisible lighting reasoning.} Although effective, our framework still struggles with inferring complex invisible objects from scarce light and shadow cues.}
    \vspace{-3mm}
    \label{fig:failure_case}
\end{figure}

\vspace{-5mm}
\paragraph{Limitation.} Our framework remains limited in inferring unseen scene objects from sparse and overlapping shadow cues.
As shown in Fig.~\ref{fig:failure_case}, it fails to reconstruct the off-frame people and sign suggested by their shadows, leaving the corresponding shadow patterns incomplete.

\noindent\textbf{Future work.} We will incorporate complex scene examples through in-context learning and post-train open-source VLMs on challenging datasets to strengthen reasoning about unseen objects and their interactions with light. Reducing token budget is also meaningful. We argue that projecting visual representation back to executable codes is a highly promising direction in this field.

\clearpage
\section*{LLM Usage Statement}
OpenAI Codex was used to assist with basic code generation, manuscript organization and language editing.
Codex was not used to formulate research ideas, design the method or experiments, or draw scientific conclusions. All AI-assisted text was manually reviewed and revised by the authors. All AI-assisted code was inspected, executed, and verified for correctness, and the resulting visualizations and preprocessed data were checked against the source data and expected outputs. The authors take full responsibility for the final content, including all text, code, results, and claims.

{
    \small
    \bibliographystyle{ieeenat_fullname}
    \bibliography{main}
}


\end{document}